\documentclass[english,runningheads]{llncs}
\usepackage[utf8]{luainputenc}
\usepackage{babel}
\usepackage{graphicx}
\usepackage{tabularx}
\usepackage{multirow}
\usepackage{graphics}
\usepackage{amsmath}
\usepackage{hyperref}
\newcommand{\mytilde}{\raise.17ex\hbox{$\scriptstyle\mathtt{\sim}$}}

\begin{document}

\title{Single- and Multi-Task Architectures for Tool Presence Detection Challenge
at M2CAI 2016}

\author{Andru P. Twinanda \inst{1} \and Didier Mutter\inst{2} \and Jacques Marescaux \inst{2} \and Michel De Mathelin \inst{1} \and Nicolas Padoy \inst{1}}

\authorrunning{ }
\titlerunning{ }

\institute{ICube, University of Strasbourg, CNRS, IHU Strasbourg, France \\ \and IRCAD, IHU Strasbourg,
University Hospital of Strasbourg, France \\}

\maketitle

\section{Introduction}

\begin{figure}[t!]
\begin{centering}
\begin{tabular}{c}
\includegraphics[width=3cm,height=2.2cm]{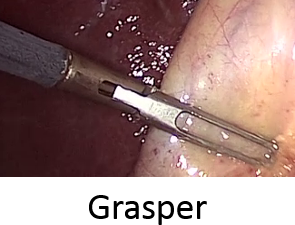} \includegraphics[width=3cm,height=2.2cm]{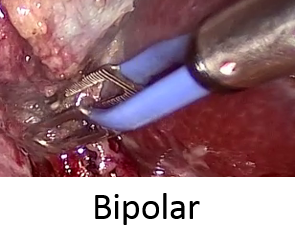}
\includegraphics[width=3cm,height=2.2cm]{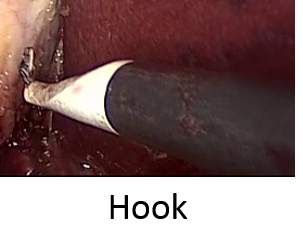} \includegraphics[width=3cm,height=2.2cm]{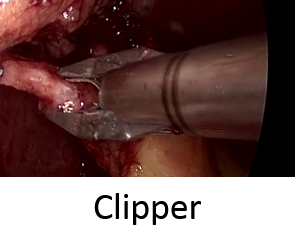}
\tabularnewline
\includegraphics[width=3cm,height=2.2cm]{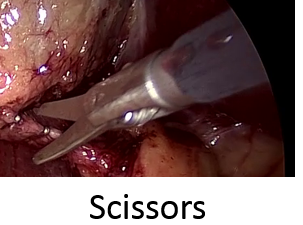} \includegraphics[width=3cm,height=2.2cm]{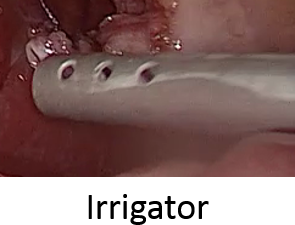}
\includegraphics[width=3cm,height=2.2cm]{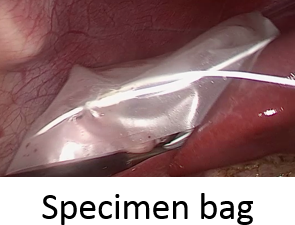}\tabularnewline
\tabularnewline
\end{tabular}
\par\end{centering}
\caption{Seven surgical tools included in the m2cai2016-tool dataset. \label{fig:defines-tools}}
\end{figure}

The tool presence detection challenge at M2CAI 2016 consists of identifying the presence/absence of seven surgical tools in the images of cholecystectomy videos. Note that the tool \textit{presence} detection task does not involve any localization of the surgical tools in the image. In Fig. \ref{fig:defines-tools},
we show the seven tools that are included in the \textit{m2cai2016-tool} dataset \cite{twinanda_tmi2016}\footnote{The dataset is available at the official web page of M2CAI 2016: \url{http://camma.u-strasbg.fr/m2cai2016/} }. The training dataset,
released on May 23 2016, consists of 10 cholecystectomy videos annotated
with the tool presence at 1 fps; while the testing dataset, released on
September 9 2016, consists of 5 videos.

Here, we propose to use deep architectures to perform the tool presence detection
task. This work is based on our previous work \cite{twinanda_tmi2016},
where we presented several network architectures to perform multiple recognition tasks on laparoscopic videos. The tasks are surgical phase recognition and tool presence detection. Ultimately, we proposed an architecture which is designed to jointly perform both tasks. In this work, we are using both single-task and multi-task networks to address the challenge.

\section{Methodology}

\begin{figure}[t!]
\begin{centering}
\begin{tabular}{cc}

\multicolumn{2}{c}{\includegraphics[width=12cm]{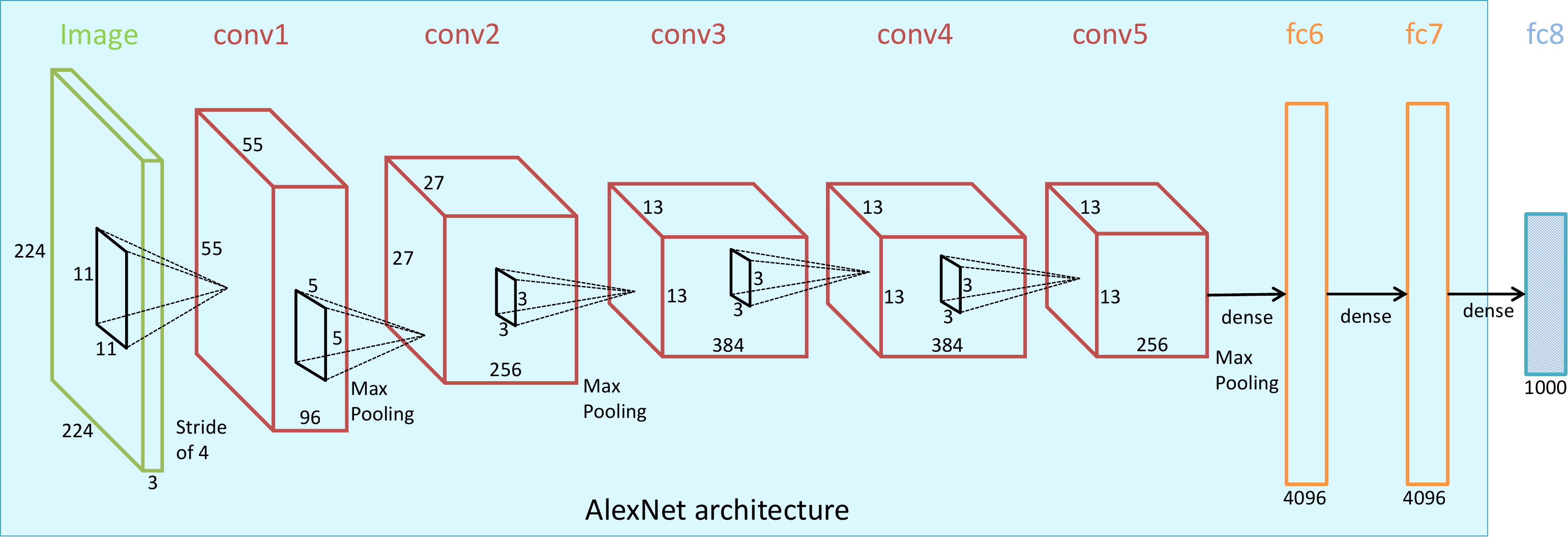}}\\
\multicolumn{2}{c}{(a)}\\

\includegraphics[height=2cm]{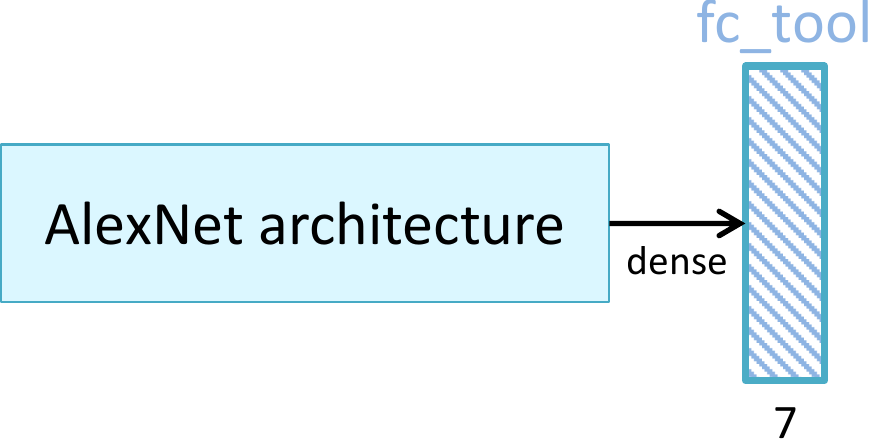} & \includegraphics[height=2cm]{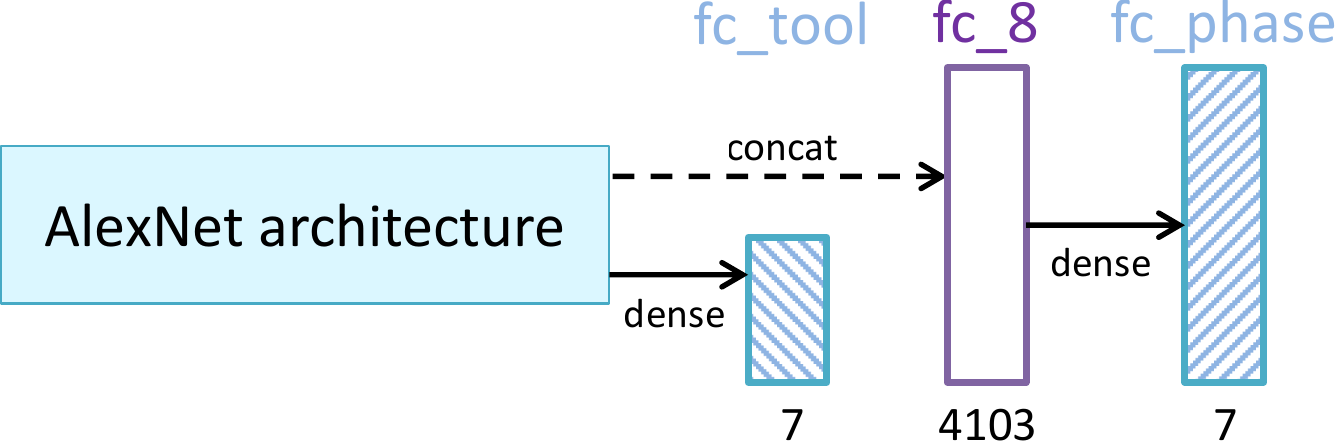} \\
(b) & (c) 
\end{tabular}
\par\end{centering}
\caption{The architectures of: (a) AlexNet, (b) ToolNet, and (c) EndoNet. \label{fig:network-architectures}}
\end{figure}

In previous work \cite{twinanda_tmi2016}, we proposed two convolutional
neural network (CNN) architectures to perform tool presence detection: ToolNet and EndoNet, shown in
Fig. \ref{fig:network-architectures}. ToolNet
is designed to solely perform the tool presence detection task, while EndoNet
is designed to jointly perform the phase recognition and tool presence
detection tasks. In \cite{twinanda_tmi2016}, it has been shown that the multi-task
network performs slightly better than the single-task counterpart. However,
the multi-task network requires both phase and tool presence annotations,
which are not available in the m2cai16-tool dataset. In Section \ref{sect:feature-comparison}, we will explain how we conduct our experiments to cope with this limitation.

The tool presence detection task is performed by using the output of the $\mathtt{fc\_tool}$ layer which contains 7 nodes (equal to the number of tools). The output of this layer corresponds to the confidences of the presence of the seven tools in the image. By applying thresholds to these confidences, we can determine the presence of the surgical tools in the image.

\section{Experimental Setup}

\subsection{Network Comparison \label{sect:feature-comparison}}

As previously stated, the EndoNet architecture is designed to perform jointly surgical phase recognition and tool presence detection while the m2cai16-tool does not contain tool binary annotations. To cope with this limitation, we are using the Cholec80 dataset \cite{twinanda_tmi2016} which contains both phase and tool binary annotations. In addition to the extra annotation, the Cholec80 dataset contains more training videos than the m2cai16-tool dataset (i.e., 40 vs. 10 training videos). Here, we will finetune multiple networks with the ToolNet and EndoNet architectures using m2cai16-tool and Cholec80.\footnote{For more information regarding the Cholec80 dataset, we refer readers to \cite{twinanda_tmi2016}.}

In summary, we are going to compare the performances of the following networks:
\begin{itemize}
\item ToolNet-m2cai16. This network is trained using the PhaseNet architecture on the m2cai16-workflow dataset;
\item ToolNet-Cholec80. This network is trained using the PhaseNet architecture on the Cholec80 dataset;
\item EndoNet-Cholec80. This network is trained using the EndoNet architecture on the EndoNet dataset.
\end{itemize}

\subsection{ToolNet and EndoNet Finetuning Parameters}

All networks are trained by fine-tuning the publicly available AlexNet network \cite{krizhevsky_nips2012} which has been pre-trained on the ImageNet dataset \cite{imagenet}. The layers that are not defined in AlexNet (i.e., $\mathtt{fc}$\_$\mathtt{tool}$ and $\mathtt{fc}$\_$\mathtt{phase}$) are initialized randomly. 
The network is fine-tuned for 50K iterations with $N_{i}=50$ images in a batch.
The learning rate is initialized at $10^{-3}$ for all layers, except
for $\mathtt{fc}$\_$\mathtt{tool}$ and $\mathtt{fc}$\_$\mathtt{phase}$,
whose learning rate is set higher at $10^{-2}$ because of their random initialization. The learning rates for all layers decrease by a factor of
$10$ for every 20K iterations. The fine-tuning process is carried
out using the Caffe framework \cite{caffe}. 


\subsection{Evaluation Metrics}

The tool presence detection challenge is evaluated using mean average precision (mAP). This metric is obtained by computing the area under the precision-recall curve. The metric is first computed for each tool and then averaged over all the tools. 

\section{Experimental Results}

We show the tool presence detection results in Table \ref{tab:tool-presence-results}. It can be seen that ToolNet-m2cai16 does not perform very well, yielding an overall mAP of 52.5. This is significantly lower compared to ToolNet-Cholec80 and EndoNet-Cholec80 which yield 73.9 and 74.2 overall mAP, respectively. ToolNet-m2cai16 yields significantly lower mAP due to the fact the m2cai16-tool dataset contains significantly fewer training videos. This result has also been observed in the previous work \cite{twinanda_tmi2016} which showed that the system performance improves as the number of training size increases. Similarly to the results in \cite{twinanda_tmi2016}, there is no significant improvement observed when the multi-task network is used for the tool presence detection task.

Looking at the mAP of ToolNet-m2cai16 for each tool, one can notice the low recognition results for scissors, clipper, and irrigator. This is most likely due to the fact that these tools are only present during short period of times in the procedure. For this reason, the training images for these tools are scarce in the dataset. In addition, these tools have appearance similarities with other tools that appear very often during the procedures (i.e., grasper). Even when a larger training dataset is used to train the networks, there is still much room for improvements in the detection results of these three tools.

\begin{table}[t!]
\begin{centering}
\begin{tabular}{|c|c|c|c|}
\hline
Tool & ToolNet-m2cai16 & ToolNet-Cholec80 & EndoNet-Cholec80 \\
\hline
\hline
Grasper & 82.2 & 86.0 & 87.0 \\
\hline
Bipolar & 50.3 & 69.1 & 68.7 \\
\hline
Hook & 89.4 & 94.2 & 93.9 \\
\hline
Scissors & 17.0 & 51.9 & 52.8 \\
\hline
Clipper & 43.6 & 63.0 & 66.5 \\
\hline
Irrigator & 12.5 & 65.1 & 63.0 \\
\hline
Specimen bag & 72.2 & 88.6 & 87.3 \\
\hline
\hline
MEAN & 52.5$\pm$30.5 & 73.9$\pm$15.7 & 74.2$\pm$15.3 \\
\hline
\end{tabular}
\par\end{centering}
\caption{Tool presence detection results. \label{tab:tool-presence-results}}
\end{table}

\section{Conclusions}

We have presented several approaches to address the tool presence detection challenge at M2CAI 2016. We proposed to use two types of CNN architectures to address the task: ToolNet and EndoNet. The former performs the tool presence detection task in a single-task manner, while the latter performs the task jointly with the phase recognition task. The results show that for tool presence detection, the multi-task architecture does not necessarily introduce high improvements to the results compared to the single-task counterpart. Instead, a significant improvement is obtained when there are more data available to train the networks. This improvement can be regarded as a call for action for other institutions to start working toward publishing more datasets into the community, so that better models could be generated to perform the task.

Here, the tool presence detection task is addressed in a frame-wise manner, i.e., there is no temporal information incorporated in the detection process. For future work, it would be interesting to see whether the temporal information plays a role in the detection process. To establish an end-to-end architecture, the temporal information can be incorporated with the usage of recurrent neural network (RNN).

\bibliographystyle{plain}
\bibliography{bibliography}

\begin{thebibliography}{1}

\bibitem{caffe}
Yangqing Jia, Evan Shelhamer, Jeff Donahue, Sergey Karayev, Jonathan Long, Ross
  Girshick, Sergio Guadarrama, and Trevor Darrell.
\newblock Caffe: Convolutional architecture for fast feature embedding.
\newblock {\em arXiv preprint arXiv:1408.5093}, 2014.

\bibitem{krizhevsky_nips2012}
Alex Krizhevsky, Ilya Sutskever, and Geoffrey~E. Hinton.
\newblock Imagenet classification with deep convolutional neural networks.
\newblock In {\em Advances in Neural Information Processing Systems 25}, pages
  1097--1105. 2012.

\bibitem{imagenet}
Olga Russakovsky, Jia Deng, Hao Su, Jonathan Krause, Sanjeev Satheesh, Sean Ma,
  Zhiheng Huang, Andrej Karpathy, Aditya Khosla, Michael Bernstein,
  Alexander~C. Berg, and Li~Fei-Fei.
\newblock {ImageNet Large Scale Visual Recognition Challenge}.
\newblock {\em IJCV}, 115(3):211--252, 2015.

\bibitem{twinanda_tmi2016}
Andru~Putra Twinanda, Sherif Shehata, Didier Mutter, Jacques Marescaux, Michel
  de~Mathelin, and Nicolas Padoy.
\newblock Endonet: A deep architecture for recognition tasks on laparoscopic
  videos.
\newblock {\em IEEE Transactions on Medical Imaging}, 2016.

\end{thebibliography}

\end{document}